\documentclass[10pt,twocolumn,letterpaper]{article}

\usepackage{setspace}
\usepackage{balance}
\usepackage{iccv}
\usepackage{times}
\usepackage{epsfig}
\usepackage{graphicx}
\usepackage{amsmath}
\usepackage{amssymb}

\usepackage{xspace}
\usepackage{graphics} 
\usepackage{epsfig} 
\usepackage{mathptmx} 
\usepackage{times} 
\usepackage{amsmath} 
\usepackage{amssymb}  
\usepackage[ruled,linesnumbered]{algorithm2e}
\usepackage{float}
\usepackage{makecell}

\usepackage{color, colortbl}
\usepackage{enumitem}
\usepackage[stable]{footmisc}
 
\usepackage[pagebackref=true,breaklinks=true,letterpaper=true,colorlinks,bookmarks=false]{hyperref}

\iccvfinalcopy 

\def\iccvPaperID{4459} 

\ificcvfinal\fi

\begin{document}
\title{The benefits of synthetic data for action categorization}

\author{Mohamad Ballout, Mohammad Tuqan, Daniel Asmar, Elie Shammas, George Sakr\\
Department of Mechanical Engineering, VRL Lab\\
American University of Beirut, Beirut, Lebanon\\
{\tt\small mab73@mail.aub.edu, m.m.tuqan@gmail.com, da20@aub.edu.lb, es34@aub.edu.lb, georges.sakr@usj.edu.lb}
}

\maketitle

\begin{abstract}
In this paper, we study the value of using synthetically produced videos as training data for neural networks used for action categorization. Motivated by the fact that texture and background of a video play little to no significant roles in optical flow, we generated simplified texture-less and background-less videos and utilized the synthetic data to train a Temporal Segment Network (TSN). The results demonstrated that augmenting TSN with simplified synthetic data improved the original network accuracy (68.5\%), achieving 71.8\% on HMDB-51 when adding 4,000 videos and  72.4\% when adding 8,000 videos. Also, training using simplified synthetic videos alone on 25 classes of UCF-101 achieved 30.71\% when trained on 2500 videos and 52.7\% when trained on 5000 videos. Finally, results showed that when reducing the number of real videos of UCF-25 to 10\%  and combining them with synthetic videos, the accuracy drops to only 85.41\%, compared to a drop to 77.4\% when no synthetic data is added.     
\end{abstract}
\section{Introduction}
In the drive towards a pervasive Internet-of-Things (IoT) society, machines will have to interact much more with humans, and to do so, they will have to understand human actions and activities. Action categorization is the process of classifying a trimmed video that contains a single action. For example, a four second video of a person biking should be classified as `biking'. On the other hand, the process of detecting the time interval of each action in a video and labeling them is called temporal action detection. In this paper, we are dealing with the problem of action categorization. 
	
\begin{figure}[t]
\begin{center}
\includegraphics[scale=0.37]{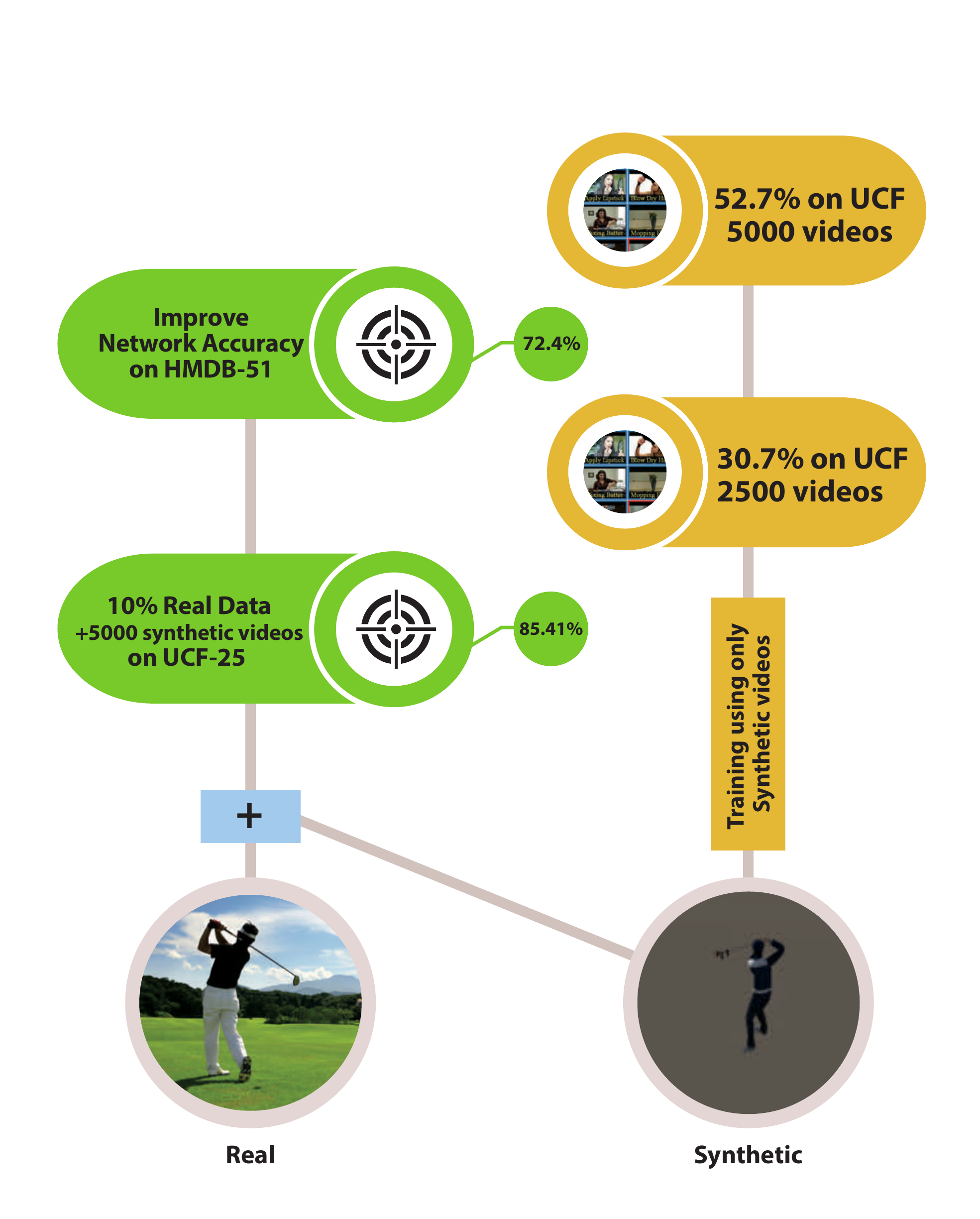} 
\caption{Synthetic data can be used to train a network from scratch or augment a pre-trained network to improve its performance. }
\label{fig:motivation}
\end{center}
\end{figure}

The state-of-the-art action categorization systems are mostly built on deep network architectures. The major challenge for these networks is the collection and annotation of a sufficient number of videos for training. The deeper the networks are, the more data is required. To give some perspective to the problem, training the 3D ResNet \cite{hara_2018_CVPR} on the UCF-101 dataset failed, despite the fact that UCF-101 contains more than thirteen thousand videos of 101 classes. In fact, to successfully train 3D ResNet, a much larger dataset (Kinetic \cite{kay_2017_Arxiv}) was required, which includes more than 300,000 videos. Manually annotating all these videos is a daunting task, and thus the need for a method that generates annotated videos in a simpler manner (Fig. \ref{fig:motivation}).

One approach to overcome the requirement for large amounts of data is to do unsupervised action localization, which does not require any annotated videos, and  aims to automatically group videos of a similar action into one class. Such systems rely on local features that can detect similar actions. Unfortunately, the results of these systems to date  \cite{jones_2014_CVPR}\cite{soomro_2017_ICCV} are not comparable to supervised systems, with accuracy values as low as  60\% on UCF-101. 

Another approach to mitigate the problem of video collection and annotation, is to rely on simulated data instead of videos of real recorded actions. Such approaches have been attempted on still images generated through graphics simulators. For example, synthetic data was used in object detection \cite{tremblay_2018_Arxiv}\cite{dwibedi_2017_ICCV}; it was also used in segmentation \cite{ros_2016_CVPR},  and also in the evaluation of optical flow solutions \cite{dosovitskiy_2015_ICCV}.  Creating simulated realistic full scenes with complete background information is difficult, and not many people have attempted it \cite{deSouza_2017_CVPR}; however, what is interesting to note is that the difficult part is mostly in the creation of the background. In fact, if one could disregard the background completely, creating actions in simulation would be relatively simple to do.   

Training deep network models using synthetic data has shown promising results in multiple computer vision applications, such as object detection \cite{tremblay_2018_Arxiv}\cite{peng_2015_ICCV}\cite{dwibedi_2017_ICCV}\cite{wu_2015_NIPS}, segmentation \cite{sankaranarayanan_2018_CVPR}\cite{ros_2016_CVPR}\cite{handa_2016_CVPR}, optical flow estimation \cite{dosovitskiy_2015_ICCV}\cite{butler_2012_ECCV}, action recognition \cite{deSouza_2017_CVPR}\cite{rahmani_2016_CVPR}\cite{rahmani_2015_CVPR}\cite{vondrick_2016_NIPS}, and pose estimation \cite{su_2015_ICCV}\cite{rogez_2016_NIPS}.  Some of these contributions are in the methods used for generating synthetic data. For instance, Dwibedi \etal \cite{dwibedi_2017_ICCV} suggested an easy yet effective way to generate synthetic images using what they call `cut and paste'. Cropped pictures where placed inside any background picture to create realistic training images. The most important point in their contribution is that images are generated quickly, and their system outperforms the existing synthesis detection approaches by 21\% when combined with real data.

Another important aspect, referred to as cross-domain generalization, is the ability of synthetic data to generalize to real world data. It would not be useful to train a system on synthetic data, and then have the trained network produce worse detection and classification on real world data. To address this problem, Tobin \etal \cite{tobin_2017_IROS} used domain randomization \cite{butler_2012_ECCV} for object detection; the concept is based on introducing random variations in the simulator in such a manner that the world, after randomization, appears different to the to AI system.  In the simulator of Tobin \etal \cite{tobin_2017_IROS}, the parameters for lighting, pose, and textures were set in a random and non-realistic manner. Their idea was to provide enough variability when training in a way that the system would then be able to generalize to the real world during testing. By training only on synthetic data, their proposed network produced compelling results on a benchmark object detection dataset. 

The method we are proposing is similar in spirit to that of De Souza \etal \cite{deSouza_2017_CVPR}; however, three main differences exist. First, the network we use is an \textit{augmented} TSN, which includes an additional channel or two for synthetic data compared to their cool-TSN, which feeds the synthetic and the real data together using mini-batches. Second, we propose using a reduced form of synthetic data (hereafter referred to as \textit{simplified synthetic data}), which leads to a better accuracy than that of \cite{deSouza_2017_CVPR}.  Third, the synthetic dataset we created and used for training is considerably smaller than that of \cite{deSouza_2017_CVPR} (8529 versus 39,982), and yet the improvement we achieve is higher than what they achieve (3.9\% versus 1\% on HMDB-51). Our experiments are more comprehensive, in which we test our augmented TSN on different inputs, sometimes using synthetic data alone, and others using a mixture of real and synthetic data. 

To the best of our knowledge, along with \cite{vondrick_2016_NIPS}, only a few papers assess the effect of training a network on synthetic videos alone, and test it on a dataset such UCF-101. Our results revealed that the more synthetic data is used to train a network, the higher the accuracy becomes. 

In this paper, we propose to evaluate the advantage of using simplified synthetic data for training neural networks for action categorization. We test our approach on the Temporal Segment Network (TSN) \cite{wang_2016_ECCV}, which learns both spatial and temporal streams separately; the temporal stream is fed with optical flow fields, and is not affected by background. For instance, if the action to be modeled is `diving from a cliff', whether the surrounding environment is simulated in the scene or not, the optical flow frames are not affected as long as the background is static. Thus, our videos generated using a physics engine, such as Unity, are background-free and no complex scene design was required. 

The contributions of this paper include the following:
\begin{itemize}
\item First, we prove the efficiency of using simplified synthetic data for action recognition, in which only optical flow data is considered. Domain randomization is applied by shaking the camera in a random fashion, as well as  changing the lighting conditions in the recorded videos.
\item Second, augmenting the vanilla TSN  \cite{wang_2016_ECCV} by including as input an additional stream of synthetic data on top of the real videos. Our proposed augmentation outperforms the vanilla TSN by a significant margin.     
\item Third, training the TSN with only the generated synthetic dataset resulted in 52.7\% accuracy when tested on the sub-dataset of UCF-101. 
\item Finally, we release a dataset of synthetic data with all the actions chosen from classes of benchmark datasets, including HMDB-51 and UCF-101. We name the datasets S-HMDB-38 and S-UCF-25, and make them publicly available for testing. \footnote{https://www.dropbox.com/s/isz6vsw6fzibbwk/datasets.zip?dl=0}
\end{itemize}	


\section{Methodology}
\label{BLR}
This section presents the proposed  methodology, including an analysis on the value of appearance data versus optical flow in action recognition, as well as the assessment of what part of synthetic data is most relevant for the sake of action categorization.  
\subsection{The Value of Appearance Data in Action Categorization}
In action recognition networks, it is common to include both spatial and temporal ConvNet streams. In TSN \cite{wang_2016_ECCV}, for example, the spatial stream takes RGB frames as input, and the temporal stream processes optical flow. Each stream is trained separately and then their scores are fused at the output layer. When tested on the UCF-101 dataset, after training on  the spatial stream alone, TSN scored an accuracy of 84.5\%,  and when trained on the temporal stream alone, it scored an accuracy of 87.2\%. Fusing both streams resulted in an improved accuracy of 92.0\%. Looking closer at these results, the relatively moderate improvement attained when adding appearance data questioned its value in action categorization. This result motivated us to use \emph{only} temporal information in our synthetic data; an idea that agrees with the results presented in \cite{jhuang_2013_ICCV}. Moreover, using only temporal information introduces substantial simplifications to the synthesis of videos, as will be seen below. 

Optical flow is the pattern of motion in the visual scene reflecting the relative motion between the object and the camera. For the sake of reproducing the action, a question arises regarding the background of the scene and its impact on the optical flow. Under ideal conditions, where the camera is fixed, and the lighting condition is consistent among all the videos within the dataset, the background does not have a significant contribution to the optical flow pattern.  However, in the absence of ideal conditions, relative motion can be detected from the pixels in the background of the scene. This happens, for instance, when a camera is held by a person who is moving or walking. This random movement or `shake' in a camera can distort the ideal conditions that are replicated in a gaming engine.

One approach to tackle this problem is domain randomization \cite{tremblay_2018_Arxiv} \cite{butler_2012_ECCV}, which is used in object recognition. In domain randomization, an object in a scene is placed out of context in order to train the network to deal with the possible variability in the scene appearance when detecting that object. In our paper,  we borrow the idea of randomization from the field object recognition, and apply it to actions by adding random changes in light intensity in the scene, and random camera movements. The variations are applied either within the vicinity of its original position to introduce the shaking effect into the camera, or by tracking the action of interest. Therefore, the synthetic videos we use constitute a combination of scenes under ideal and non-ideal conditions, either by introducing distortion in light intensity, camera position, or a combination of both. 
\subsection{Synthetic Data}\label{sec.syn_data}
\begin{figure*}[t!]
\begin{center}
\includegraphics[scale=0.8]{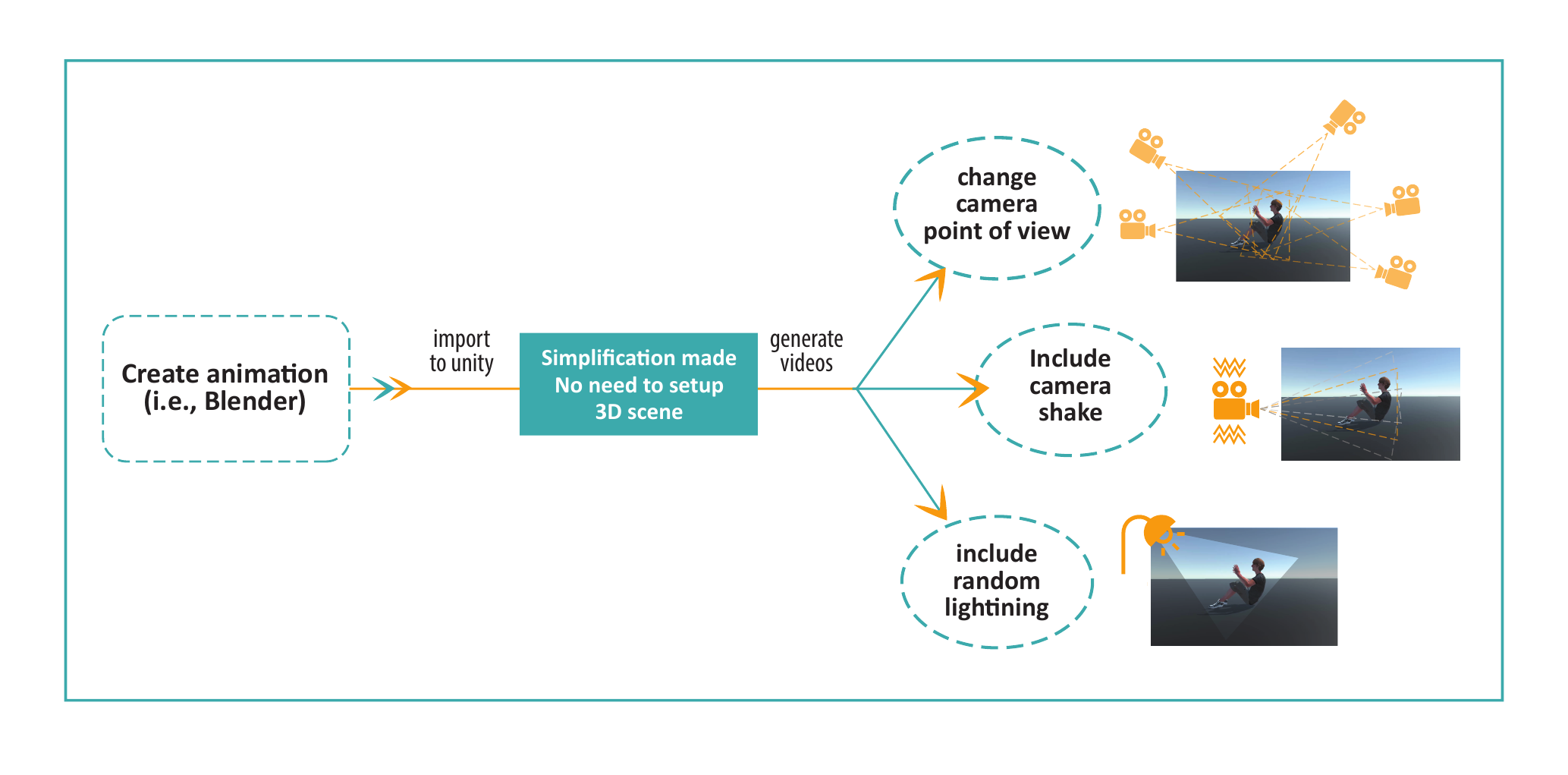} 
\caption{The simplified data were generated using Unity without setting up a 3D scene}
\label{fig:data generator}
\end{center}
\end{figure*}

Creating realistic backgrounds is one of the main challenges one faces when generating synthetic data. In an outdoor environment, this challenge is even more difficult, requiring graphics experts capable of reproducing computer models of natural objects, such as trees and rocks. It would seem that in the case where background information is critical, the required computer effort could be better spent collecting videos of real scenes. On the other hand, if one could completely disregard the background and create simple videos of foreground alone, synthetic data creation becomes considerably simpler (see Fig. \ref{fig:data generator}).  

We investigate the significance of appearance on network performance for videos of human actions on simulated data we created,  as well as on the benchmark datasets HMDB-51 and UCF-101.  The synthetic actions were downloaded from Mixamo.com, or from the Unity asset store.  To test for the effect of background, various backgrounds and scenes were downloaded from the Unity asset store. For the HMDB-51 classes, 38 were reconstructed; whereas for the 101 UCF, 25 classes were reconstructed.  Table \ref{tab:videos} shows the number of videos and classes for actions corresponding to those found in the benchmark datasets. It is worth noting that we only synthesized videos for 38 out of the 51 HMDB classes, and 25 our of the 101 UCF classes, since other classes were not available to download. 
\begin{small}
\begin{table}[b]
\caption{Number of videos created of objects corresponding to those found in benchmark datasets: background videos stands for the videos that were made using a 3D setup.}
\begin{center}
\begin{tabular}{|c|c|c|c|}
\hline
\textbf{Dataset}& \makecell{Classes \\(\#)} & \makecell{Background \\ Videos  (\#)}& \makecell{ Background-less  \\ Videos (\#)}  \\
\hline
HMDB-51 & 38& 3817& 8528\\
\hline
UCF-101&25 & -- & 5514 \\
\hline
\end{tabular}
\label{tab:videos}
\end{center}
\end{table}
\end{small}

\begin{figure}[t]
\begin{center}
\includegraphics[scale=0.45]{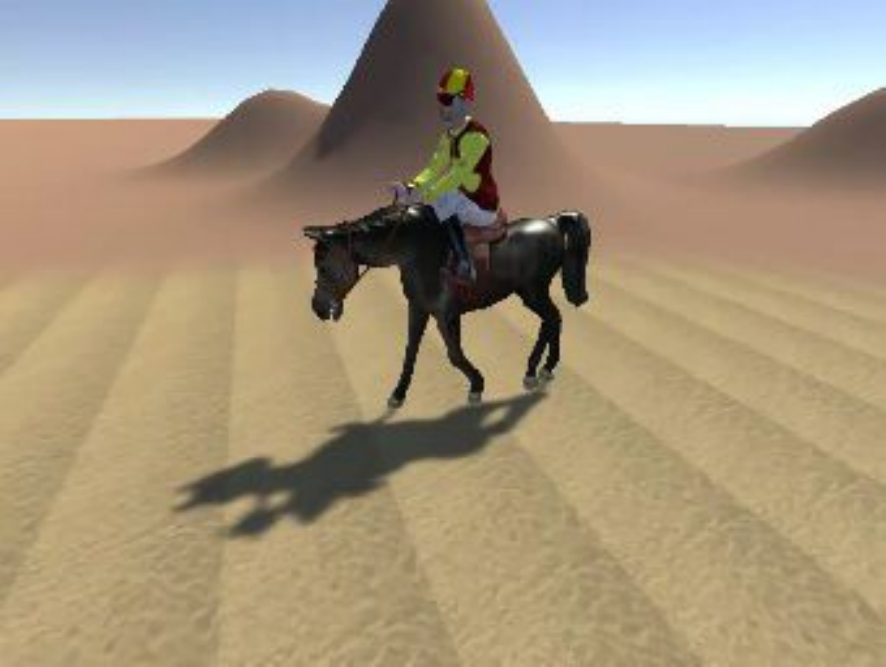} \includegraphics[scale=0.45]{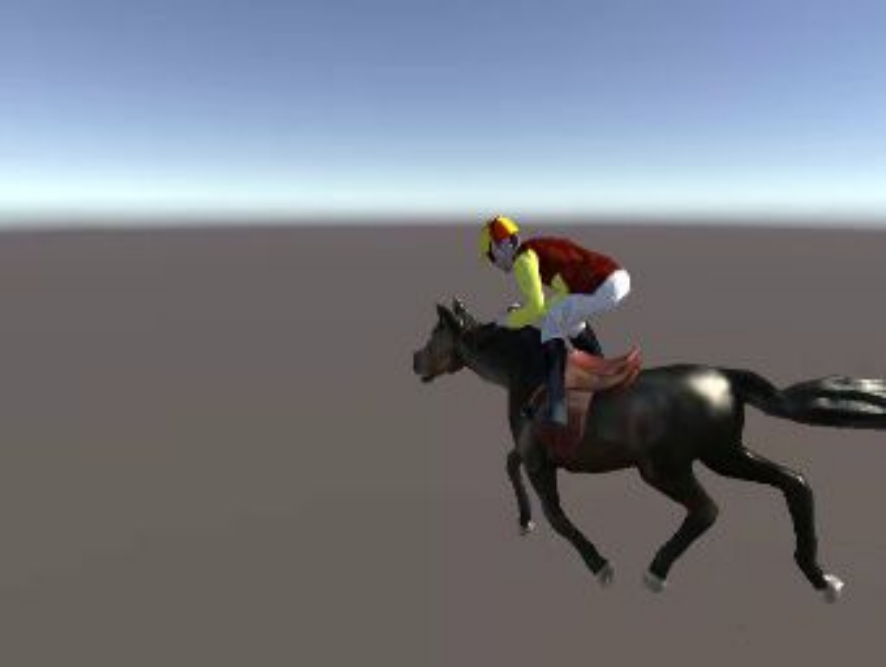}\\
\includegraphics[scale=0.45]{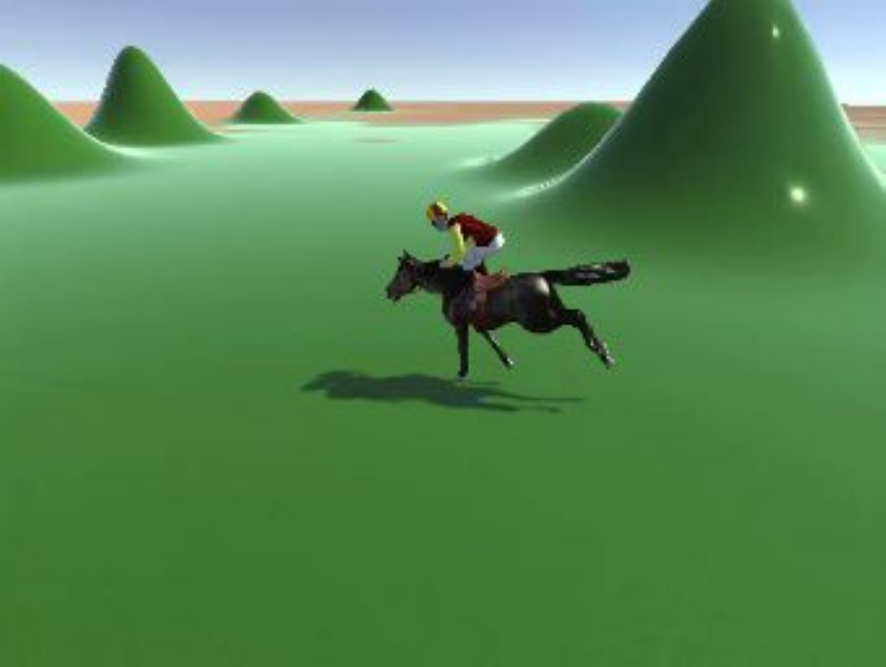} \includegraphics[scale=0.45]{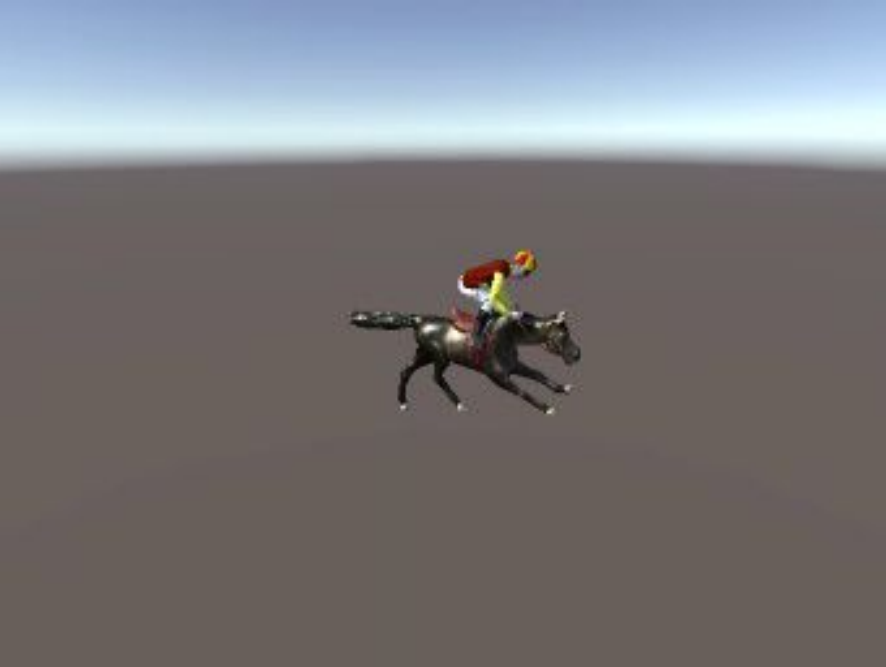}\\
\includegraphics[scale=0.45]{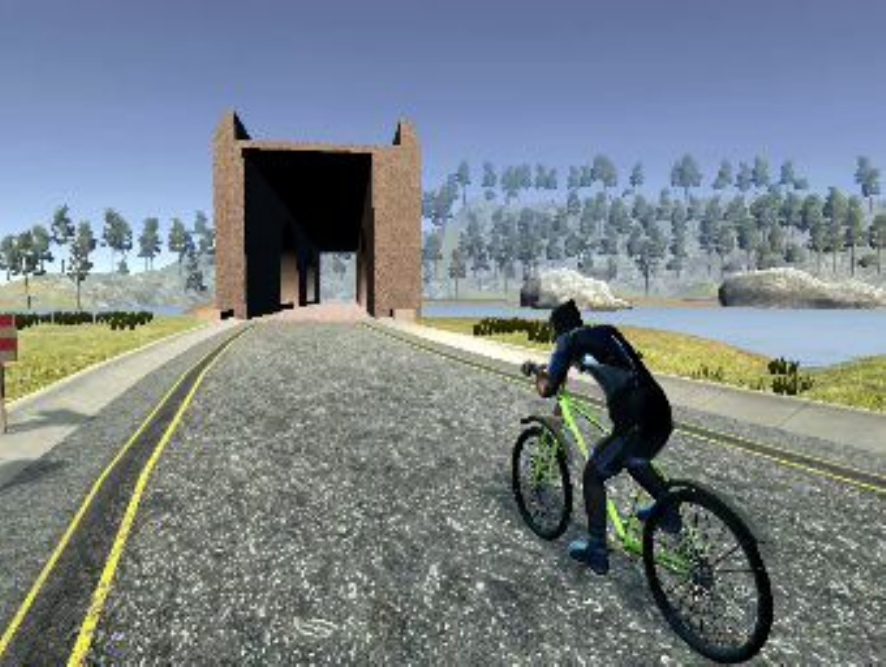} \includegraphics[scale=0.45]{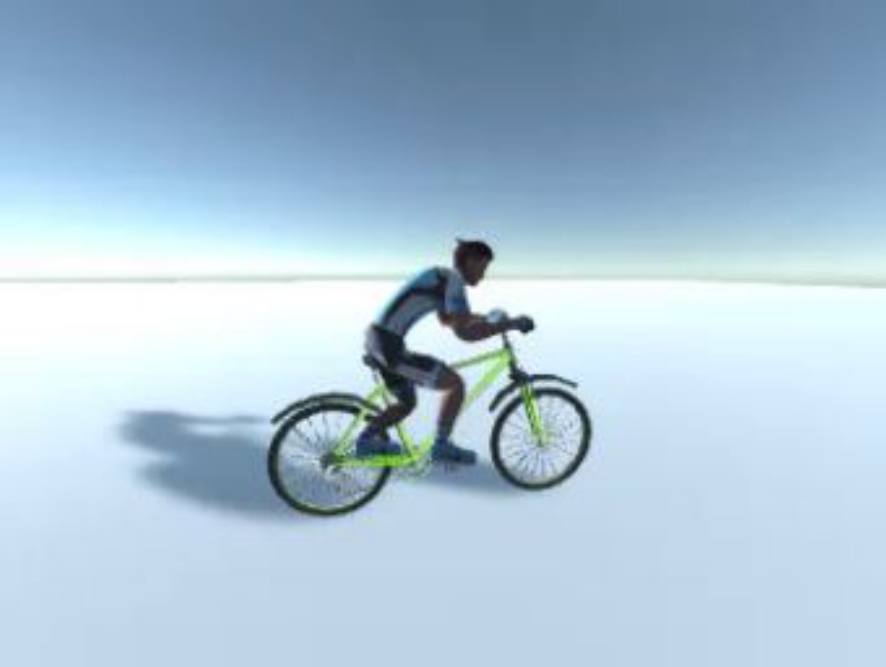}\\
\includegraphics[scale=0.45]{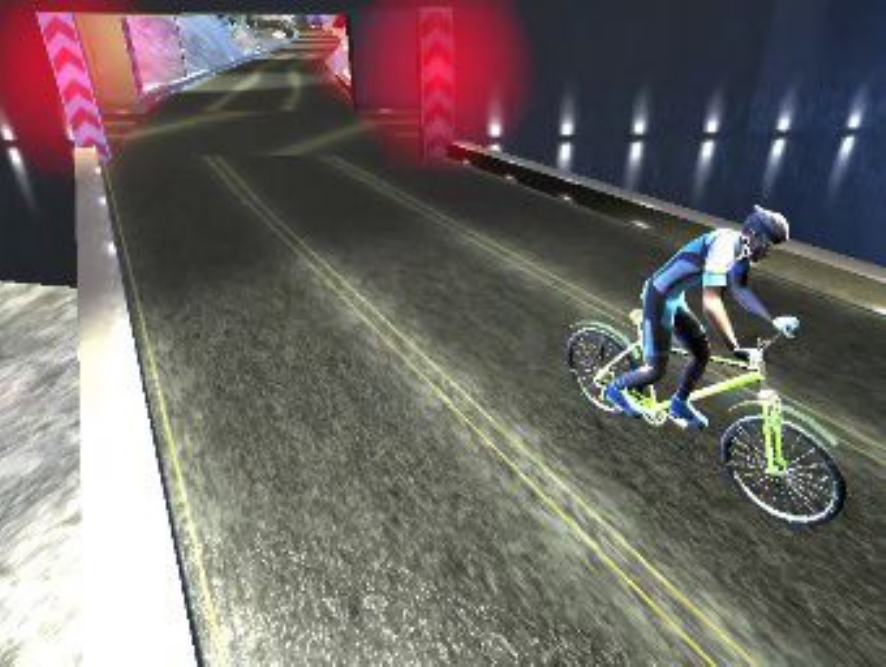} \includegraphics[scale=0.45]{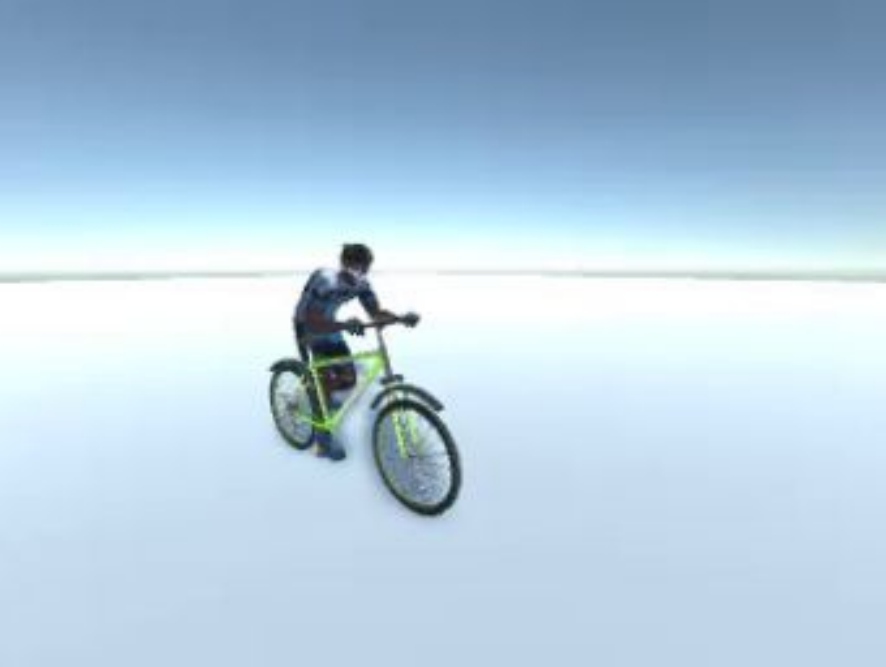}\\
\caption{Examples of videos created with background (left column) and simplified videos (right column}
\label{fig:synthetic}
\end{center}
\end{figure}

In each dataset, different characters are used to do the same action; and for each action, there are multiple animations that differ from each other in the way of doing the action. For example, if the action is `riding a bike', there are several ways of riding it: it could be ridden at a quick or slow speed, it could be ridden standing up or sitting down (see Fig. \ref{fig:synthetic}).   Also, for further generalization, some of the actions were done under different lighting conditions such as under dark, shadow, or bright lighting conditions. 
When removing appearance and relying on optical flow alone, none of these nuances are relevant any longer. Optical flow is not affected by the lighting, nor by the color and texture of the garment of a character. This fact allowed us to use the same character for most of the videos where appearance is disregarded. 

As a result, in this paper we suggest using only the optical flow of synthetic data (which is simple to obtain), and disregard the background scene information (which can be difficult to obtain).  For example, in producing the `climb stair' action, it is enough to only reproduce the action of climbing the stairs without the need to place the stairs in the scene.  For the `pitching' action, it is sufficient to simulate the act of pitching a ball without having to simulate a ball in the scene, as its contribution to the optical flow is negligible. In other words, the only factor contributing in the synthesis of the scene is the action of the character itself. 

Each action includes between 200-250 videos, requiring between 60-75 minutes in total to generate. Similar to real videos, the generated videos are between two to six seconds. Camera acquisition is set to thirty frames per second in all of the generated videos. The aspect ratios of the videos are close to those of the UCF-101 videos (320x240 pixels).  To complete the dataset, each action is reproduced from different camera viewpoints and environment conditions (light source intensity and camera position). Note however, that not all the actions are one-person synthetic. Some actions such as `salsa spin' or `boxing', require including the second person in order to correctly interpret the overall meaning of the action (two-people synthetic). 
\begin{figure*}[!t]
\begin{center}
\includegraphics[scale=0.4]{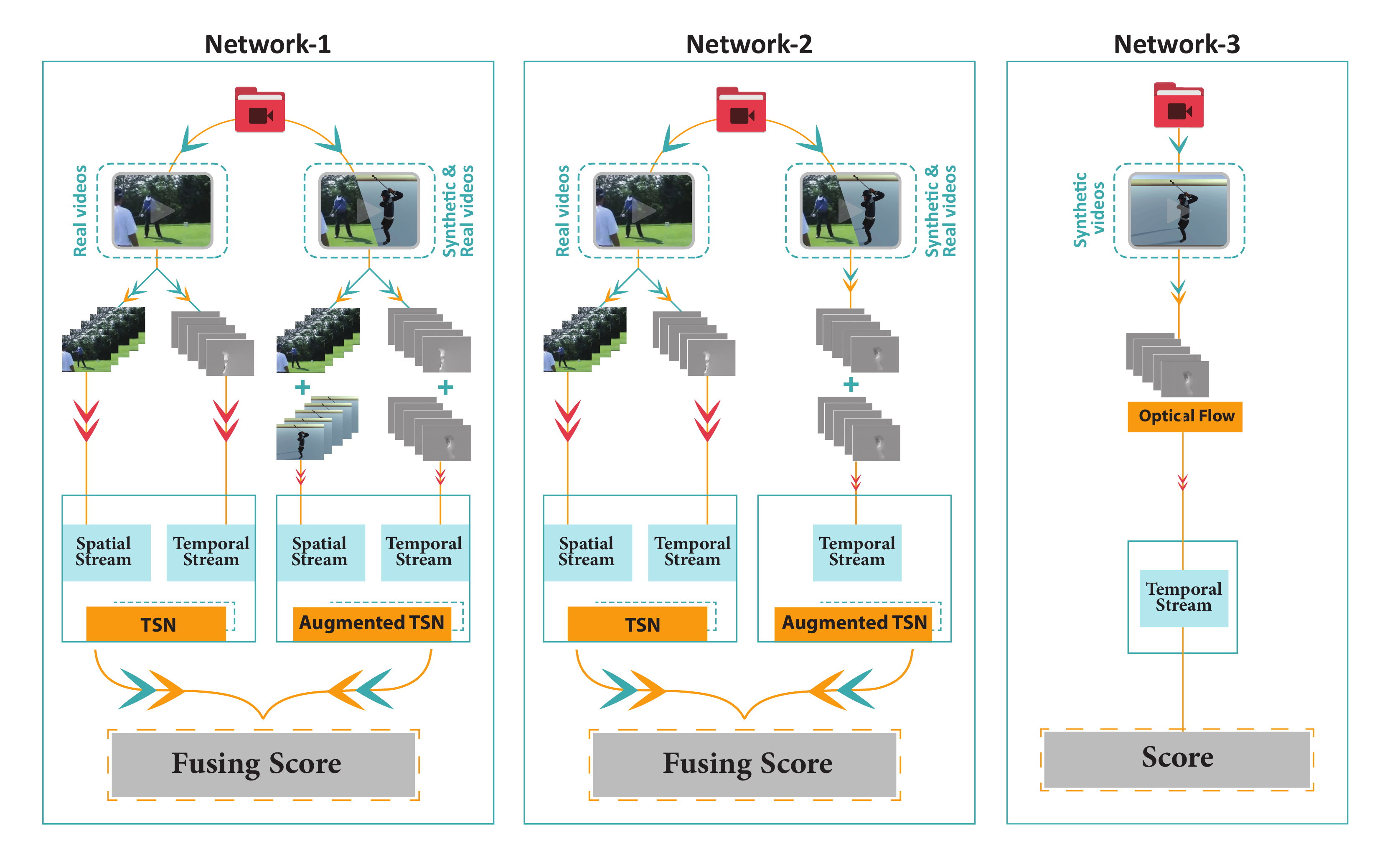} 
\caption{Networks tested, Augmented TSN with 4 and 3 streams, and Optical Flow TSN}
\label{fig:networks}
\end{center}
\end{figure*}

The results of this comparison were quite surprising: training using optical flow alone on both the real and synthetic datasets produced classifications results superior to those in which appearance data was also included.  As a result, it was decided that in simulations, we would only vary the camera viewpoint while performing the various actions.  Also, purposely shaking the camera in simulations resulted in conditions that are close to what is experienced in the real world.  

\section{Experiments and Results}  
\label{results}
Several experiments were performed to assess the effectiveness of using synthetic data in  training of action recognition networks.  In what follows, we first present the datasets we used for the experiments, then discuss the proposed networks, and finally discuss the results of our experiments.

\subsection{Datasets} \label{sec_dataset}
Five different datasets were used for the proposed experiments, two of which comprise videos of real actions, including UCF-101 and HMDB-51, and three that are synthetic, which we simulated:
\begin{itemize} 
\item UCF-101: a dataset of human actions from videos in the wild, containing 13320 videos distributed in 101 action categories
\item HMDB-51: a video database for human actions, mostly extracted from movies and web videos. It consists of 6766 videos from 51 different action categories. 
\item Synthetic-appearance (HMDB-38):  approximately 4000 synthetic videos were created including background, chosen from 38 classes of real HMDB-51. Examples are shown in the left column of Fig. \ref{fig:synthetic} 
\item Synthetic-simplified (HMDB-38): approximately 8000 synthetic videos were created using no-background, chosen from 38 classes of real HMDB-51. Examples of the videos are shown in the right column of Fig. \ref{fig:synthetic}
\item Synthetic-simplified (UCF-25): approximately 5000 synthetic videos were created using no-background, from 25 classes of the real UCF-101.
\end{itemize}

\subsection{Networks tested}
To investigate the potential of using simplified synthetic data for training, we tested four different networks (Fig. \ref{fig:networks}), including TSN, Augmented TSN (Network-1 and Network-2), and one-flow TSN (Network-3).

\paragraph{TSN} Temporal Segment Networks \cite{wang_2016_ECCV} are an upgraded version of the two stream convolutional network \cite{simonyan_2014_NIPS}. TSN uses a deeper network than the two stream network, and it benefits from a 2D-CNN model trained on ImageNet \cite{deng_2009_CVPR}. The 2D-CNN could be Inception, 2D-ResNet, or Batch-Normalized-Inception. Another advantage of TSN versus a classic two stream convolution network, is its ability to model long-range temporal structures by extracting sparse snippets of the video, instead of using consecutive frames that are highly redundant.
\paragraph{Augmented-TSN (Network-1).} This network is based on TSN, with two additional streams. The first addition is the spatial stream that takes as input synthesized + real appearance data.  The second additional stream is the temporal one, which takes as input the optical flow of both the synthetic and the real datasets. Fusing the score was done by giving the real+synthetic flow a weight of 2.0, real flow 1.0, real RGB 1.0, and synthetic+real RGB 0.5.
\paragraph{S-TSN (Network-2).} Synthetic-Simplified TSN is based also based on the TSN model, with an addition stream that trains extracted optical flow for both real and simplified synthetic data. Fusing the score was done by giving all of the streams the same weights.
\paragraph{One-flow-TSN (Network-3).} This model uses only the temporal stream of TSN. It takes as an input the extracted optical flow  of the Synthetic-UCF-25 dataset and it is tested on real videos from UCF-25. The intent of this experiment is to prove that synthetic data alone is sufficient to train a network that is later tested on real data. 
\subsection{Analysis of the Results}
All of the network parameters were tuned to those of the original TSN \cite{wang_2016_ECCV}, except for the drop-out ratio, which gave a better accuracy when set to 0.8 for the temporal stream. The network weights were initialized with pre-trained models from ImageNet using BN-Inception \cite{ioffe_2015_Arxiv} as the ConvNet architecture. Also, mini-batch stochastic gradient descent was used with a batch-size set to 128 with a momentum of 0.9. The number of iterations was adopted from the TSN for both spatial and temporal streams. The optical flow frames were extracted using the code provided in the TSN framework.  Finally, we followed the traditional evaluation on the three training/testing splits for all of the experiments.  
\subsubsection{The Effect of Background Removal}\label{sec.eff_back}
In this experiment we tested the effect of using quality versus quantity videos. For Network-1, we added 4000 synthetic videos with an appropriate background setup. For Network-2, we added 4000 simplified videos and in another instance, we added 8000 simplified videos. What we observed ( Table \ref{tab:net2vsnet1}) was that adding 8000 videos to Network-2 produced slightly better results than Network-1. Even though Network-1 includes an additional stream compared to Network-2, the latter preformed better. We concluded from this experiment that a large number of simplified videos can outperform videos with background included.
\begin{small}
\begin{table}[htbp]
\caption{Network 2 versus Network 1: note that domain randomization is effective in simulating real outdoor data}
\begin{center}
\begin{tabular}{|c|c|}
\hline
\textbf{Method}& HMDB-51 \\
\hline
Network-1 with 4000 & \\ background synthetic videos & 72.3 \% \\
\hline
Network-2 with 4000 & \\ simplified synthetic videos & 71.8 \%  \\
\hline
Network-2 with 8000 & \\ simplified synthetic videos & 72.4 \%  \\
\hline
\end{tabular}
\label{tab:net2vsnet1}
\end{center}
\end{table}
\end{small}
\subsubsection{Effect of Adding Synthetic Data to Training}
This experiment was divided into two parts: the first one performed on a sub-dataset of HMDB-51 and UCF-101. As mentioned in Section~\ref{sec_dataset}, we reconstructed (synthetically) 38 classes of the HMDB-51 and 25 classes of the UCF-101. Thus, the training and the testing in this part is performed on 38 classes of HMDB-51 and 25 classes UCF-101. The purpose of this experiment is to study the effect of adding synthetic data to
all of the dataset classes. First, we reproduced the TSN results for those sub-datasets using the original real videos. Next, we added half of the generated synthetic videos (2500 for UCF-25, and 4000 for HMDB-38) to a third stream using Network-2. Finally, we added all of the generated videos (5000 for UCF-25 and 8000 for HMDB-38) to the third stream in Network-2. Table \ref{tab:subdataset} clearly shows that adding simplified synthetic videos improved the performance of TSN. It also shows that the more synthetic videos are added the better the network performs. 

The second part of the experiment is done on the entire dataset, even though some of the classes are not augmented with synthetic data. In this part, we compared the performance of Network-2 versus state-of-the-art methods, on both the HMDB-51 and UCF-101.

\begin{small}
\begin{table}[htbp]
\caption{The effect of combining real and simplified synthetic videos on sub-datasets. }

\begin{center}
\begin{tabular}{|c|c|c|}
\hline
\textbf{Dataset} & HMDB-38 & UCF-25 \\
\hline
Real only & 71.8 & 96.66\\
\hline
Half Synthetic & & \\ Videos+Real & 73.66 & 97.5 \\
\hline
All Synthetic & & \\ Videos+Real & 74.62 & 97.8 \\

\hline
\end{tabular}
\label{tab:subdataset}
\end{center}
\end{table}
\end{small}

\paragraph{HMDB-51 Dataset.}  First, we compare our Network-2 to Cool-TSN \cite{deSouza_2017_CVPR}. Cool-TSN includes 39,982 appearance videos for 35 action classes, where 21 of them are common with HMDB-51. In our Network-2, we generate only 8000 videos, with actions similar to the 38 classes of HMDB-51. In the third stream, the synthetic videos were added to the real videos in the 38 classes, while the remaining 13 classes had only real videos. Results show that Network-2 outperformed Cool-TSN by 2.9 \%.  As can be seen in Table \ref{tab:exp2}, the Network-2 also outperformed all of the state of the art systems, except I3D and OFF. However, we must note that I3D \cite{carreira_2017_CVPR} had to bfe pre-trained on 300,000 videos from the Kinetics dataset. 
\begin{small}
\begin{table}[htbp]
\caption{Benchmarking Network-2 versus state-of-the-art networks}
\begin{center}
\begin{tabular}{|c|c|c|}
\hline
\textbf{Method}& \makecell{UCF-101 \\(\%)}  & \makecell{HMDB-51 \\(\%)}  \\
\hline
iDT+FV \cite{wang_2013_ICCV} & 84.8 & 57.2 \\
\hline
Tow-stream \cite{simonyan_2014_NIPS} & 88.0 & 59.4 \\
\hline
Two-stream LSTM \cite{yue_2015_CVPR} & 88.6 & - \\
\hline
L$^2$STM \cite{sun_2017_ICCV} & 93.6 &  66.2 \\
\hline
TSN-2M \cite{wang_2016_ECCV} & 94.0 & 68.5 \\
\hline
TSN-3M \cite{wang_2016_ECCV} & 94.2 & 69.5 \\
\hline
Cool-TSN \cite{deSouza_2017_CVPR} & 94.2 &  69.5\\
\hline
3D-ResNet-101 \cite{hara_2018_CVPR} & 94.5 & 70.2 \\
\hline
Two-stream MiCT \cite{zhou_2018_CVPR} & 94.7 &  70.5\\
\hline
CoViar \cite{wu_2018_CVPR} & 94.9 & 70.2 \\
\hline
OFF \cite{sun_2018_CVPR} & 96.0 & 74.2 \\
\hline
Two-stream I3D \cite{carreira_2017_CVPR} & 98.0 & 80.7  \\
\hline
Network-2 & & \\ half of the generated & & \\ videos (Ours) & 94.4  &  71.8\\
\hline
Network-2 & & \\ all of the generated & & \\ videos(Ours) & 94.6  &  72.4\\
\hline
\end{tabular}
\label{tab:exp2}
\end{center}
\end{table}
\end{small}
On the other hand, although OFF uses 5 streams while Network-2 uses 3 streams, it outperforms our proposed system by only 1.8 \%. 

\paragraph{UCF-101 dataset.} Similar to HMDB-51, we created a sub-UCF-101 synthetic, including approximately 5000 videos,  representing 25 classes of the actions. Using the same procedure as that performed on the HMDB-51, we added the 2500 videos first and then added 5000 videos to 25 classes of UCF-101. Since we reconstructed only 25\% (25 out of 101) of the UCF-101 classes compared to 75\% ( 38 of 51) of the HMDB-51, here the improvement was lower. Network-2 improved 0.6\% above the vanilla TSN when 5000 videos were added to get a final score of 94.6 \%. 

\subsubsection{The Effect of Decreasing the Number of Real Videos}
In this section, we investigate the ability of simplified synthetic videos to replace real videos. Three tests were performed: (1) reducing the amount of videos by half, (2) keeping only 10 \% of the real videos, and (3) having no real videos at all in the training sets. 

In the first experiment, half of the real videos were randomly removed from the training splits, and the performance of the network was evaluated with and without synthetic data. In the second experiment, only 10 \% of the real data was kept and the same evaluation was repeated. Finally, in the third experiment, training was performed on only synthetic data, and testing on real data. The first two experiments were done using Network-2 while the last experiment wass done using Network-3. Network-3 in Fig. \ref{fig:networks} includes only the optical flow stream. Since there were some actions from the real datasets that could not be synthesized, during testing we limited the actions from UCF-101 to those that were produced synthetically. For UCF-101, 25 out of 101 classes were used. 
Table \ref{tab:unity} shows that the accuracy of the network did not drop by much when half of the real videos were removed. However, when only 10\% of the videos were kept, the accuracy dropped to 77.14 \% with real videos, while it stayed relatively high 85.41\% with the additional 5000 synthetic videos.
\begin{small}
\begin{table}[htbp]
\caption{The effect of decreasing the number of real videos on the accuracy of UCF-25 while adding 2500 and 5000 synthetic videos }
\begin{center}
\begin{tabular}{|c|c|c|c|}
\hline
\textbf{UCF-25} & Real & Real + 2500 & Real + 5000 \\
\hline
100\% Real & 96.66 & 97.5 & 97.8 \\
\hline
50\%Real & 96.34 & 97.02 & 97.7  \\
\hline
10\%Real & 77.4 & 81.69 & 85.41 \\
\hline
0\% Real & - & 30.71 & 52.7 \\
\hline
\end{tabular}
\label{tab:unity}
\end{center}
\end{table}
\end{small}

\begin{figure*}[!t]
\begin{center}
\includegraphics[scale=0.72]{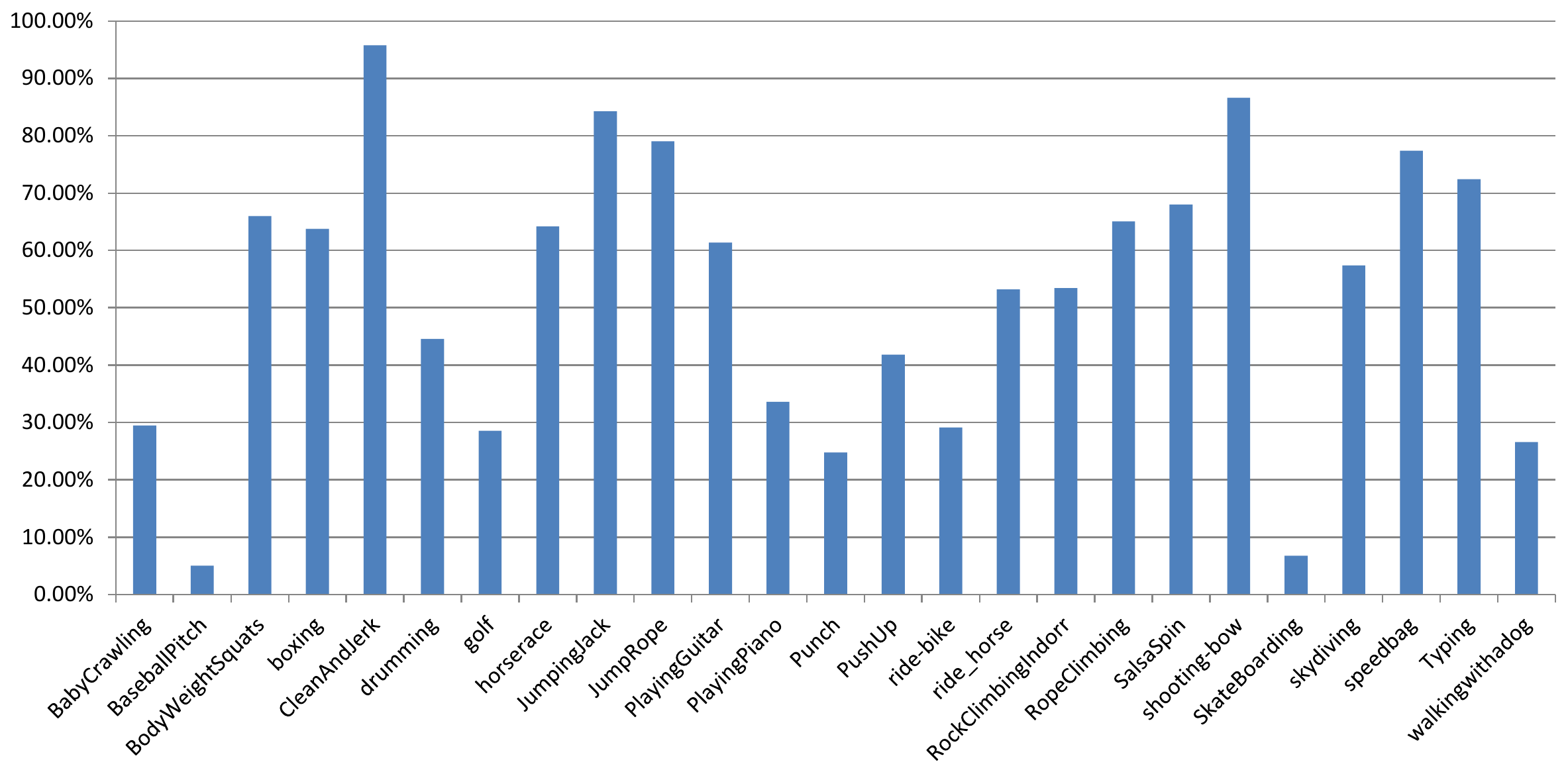} 
\caption{The accuracy for individual classes using synthetic data alone}
\label{fig:individual}
\end{center}
\end{figure*}

Finally, when training with 5000 synthetic videos, the accuracy dropped to 52.7\% while training on around 2500 synthetic videos scored 30.71\%. These results are comparable to those by \cite{vondrick_2016_NIPS}, who got 52.1\% when trained only on their generated synthetic data. The results we put forward demonstrate the potential of training action recognition networks using simplified videos, especially that some of the trained classes, such as `CleanAndJerk' achieved an accuracy above 90\%, as shown in Figure \ref{fig:individual}. On the other hand, other classes (such as `BaseBallPitch' and `Skateboarding') scored as low as 5\%, suggesting the inability of the simulator in accurately re-creating those classes. 


\section{Conclusion}\label{conc}
In this paper, we analyzed the effectiveness of simplified synthetic data in the training of deep networks for the sake of action categorization. We validated that optical flow information was sufficient to train a network, and that appearance information could be disregarded. The caveat here is that the proposed actions do not require background interaction to differentiate between two different actions (\eg, swimming vs flying). We also tested using synthesized data for training under two different scenarios: the first using synthesized data to augment TSN with an additional stream, and the second using only synthesized data to train the network from scratch. Both scenarios revealed good results, where in all cases we obtained notable improvements for TSN on both UCF-101 and HMDB-51. 
We are currently working on creating a large synthetic dataset that includes over 200 k videos in order to compare its effectiveness to that of a  dataset of real actions of comparable size, such as  Kinetics.


{\small
\balance
\bibliography{ballout_2019_ICCV.bib}
}
\end{document}